\documentclass[runningheads]{llncs}

\usepackage{times}
\usepackage{epsfig}
\usepackage{bbm}
\usepackage{array}
\usepackage{verbatim}
\usepackage{textcomp}
\usepackage[font=small,skip=0pt]{caption}
\usepackage{graphicx}
\usepackage{amsmath,amssymb} % define this before the line numbering.
\usepackage{color}
\usepackage{mathtools}
\usepackage{subfig}
\usepackage[width=122mm,left=12mm,paperwidth=146mm,height=193mm,top=12mm,paperheight=217mm]{geometry}

\begin{document}
% \linenumbers
\pagestyle{headings}
\mainmatter
\title{Riemannian Metric Learning for Symmetric Positive Definite Matrices} % Replace with your title

\titlerunning{Riemannian Metric Learning for SPD Matrices}

\authorrunning{Raviteja Vemulapalli, David W. Jacobs}

\author{Raviteja Vemulapalli, David W. Jacobs}
\institute{University of Maryland, College Park}

\maketitle

\begin{abstract}
Over the past few years, symmetric positive definite matrices (SPD) have been receiving considerable attention from computer vision community. Though various distance measures have been proposed in the past for comparing SPD matrices, the two most widely-used measures are affine-invariant distance and log-Euclidean distance. This is because these two measures are true geodesic distances induced by Riemannian geometry. In this work, we focus on the log-Euclidean Riemannian geometry and propose a data-driven approach for learning Riemannian metrics/geodesic distances for SPD matrices. We show that the geodesic distance learned using the proposed approach performs better than various existing distance measures when evaluated on face matching and clustering tasks.
\end{abstract}
\section*{Notations}
\begin{itemize}
\item $I$ denotes the identity matrix of appropriate size.\\[-5pt]
\item $\left\langle\ ,\ \right\rangle$ denotes an inner product.\\[-5pt]
\item $S_n$ denotes the set of $n \times n$ symmetric matrices.\\[-5pt]
\item $S_n^{++}$ denotes the set of $n \times n$ symmetric positive definite matrices.\\[-5pt]
\item $\mathcal{T}_p\mathcal{M}$ denotes the tangent space to the manifold $\mathcal{M}$ at the point $p \in \mathcal{M}$.\\[-5pt]
\item $\|\ \|_F$ denotes the matrix Frobenius norm.\\[-5pt]
\item Chol(P) denotes the lower triangular matrix obtained from the Cholesky decomposition of a matrix P.\\[-5pt]
\item exp() and log() denote matrix exponential and logarithm respectively.\\[-5pt]
\item $\frac{\partial}{\partial x}$ and $\frac{\partial^2}{\partial x^2}$ represent partial derivatives.\\[-5pt]
\end{itemize}
\newpage
\section{Introduction}
Many computer vision applications involve features that obey specific constraints. Such features often lie in non-Euclidean spaces, where the underlying distance metric is not the regular $\ell_2$ norm. For instance, popular features like shapes, rotation matrices, linear subspaces, symmetric positive definite (SPD) matrices, etc. are known to lie on Riemannian manifolds. In such cases, one needs to develop inference techniques that make use of the underlying manifold structure.

Over the past few years, manifolds have been receiving considerable attention from the computer vision community. In this work, we focus our attention on the set of SPD matrices. Examples of SPD matrices in computer vision include diffusion tensors [1], structure tensors [2] and covariance region descriptors [3]. Diffusion tensors arise naturally in medical imaging [1]. In diffusion tensor magnetic resonance imaging (DT-MRI), water diffusion in tissues is represented by a diffusion tensor characterizing the anisotropy within the tissue. In optical flow estimation and motion segmentation, structure tensors are often employed to encode important image features, such as texture and motion [2]. Covariance region descriptors are used in texture classification [3], object detection [4], object tracking, action recognition and face recognition [5]. There are several advantages of using covariance matrices as region descriptors. Covariance matrices provide a natural way of fusing multiple features which might be correlated. The diagonal entries of a covariance matrix represent the variance of individual features and the non-diagonal entries represent the cross correlations. The noise corrupting individual samples is largely filtered out with an averaging filter during covariance computation. Covariance matrices are low dimensional compared to joint feature histograms. Covariance matrices do not have any information regarding the ordering and the number of points. This implies a certain level of scale and rotation invariance over the regions in different images.

Various distance measures have been proposed in the literature for the comparison of SPD matrices. Among them, the two most widely-used distance measures are the affine-invariant distance [1] and the log-Frobenius distance [6] (also referred to as log-Euclidean distance in the literature). The main reason for their popularity is that they are geodesic distances induced by Riemannian metrics.

The log-Euclidean framework [6] proposed by Arsigny \textit{et.\ al.}\ defines a class of Riemannian metrics, rather than a single metric, called log-Euclidean Riemannian metrics. According to this framework, any inner product $\left\langle\ ,\ \right\rangle$ defined on $\mathcal{T}_IS_n^{++} = \{\text{log}(P)\ |\ P \in S_n^{++} \} = S_n$ extended to $S_n^{++}$ by left- or right- multiplication is a bi-invariant Riemannian metric. Equipped with this bi-invariant metric, the space of SPD matrices is a flat Riemannian space and the geodesic distance corresponding to this bi-invariant Riemannian metric is equal to the distance induced by $\left\langle\ ,\ \right\rangle$ in $\mathcal{T}_IS_n^{++}$. Surprisingly, this remarkable result has not been used by the computer vision community. Since $\mathcal{T}_IS_n^{++} = S_n$ is a vector space, this result allows us to learn log-Euclidean Riemannian metrics and corresponding log-Euclidean geodesic distances from the data by using Mahalanobis distance learning techniques like information-theoretic metric learning (ITML) [7] and large margin nearest neighbor distance learning [8] in $\mathcal{T}_IS_n^{++}$. In this work, we explore this idea of data driven Riemannian metrics/geodesic distances for the set of SPD matrices. For learning Mahalanobis distances in $\mathcal{T}_IS_n^{++}$ we use the ITML technique.

\textbf{Organization:} In section 2, we provide a brief overview of various distance measures used in the literature to compare SPD matrices.  We briefly explain the ITML technique in section 3 and present our approach for learning log-Euclidean Riemannian metrics/log-Euclidean geodesic distances from the data in section 4. We provide some experimental results in section 5 and conclude the paper in section 6.
\section{Distances to compare SPD matrices}
Various distance measures have been used in the literature to compare SPD matrices. Each distance has been derived from different geometrical, statistical or information-theoretic considerations. Though many of these distances try to capture the non-linearity of SPD matrices, not all of them are geodesic distances induced by Riemannian metrics. Tables 1 and 2 summarize these distances and their properties. Among them, the log-Frobenius distance[6] and the affine-invariant distance[1] are the most popular ones. 
\section{Mahalanobis distance learning using ITML}
Information theoretic metric learning [7] is a technique for learning Mahalanobis distance functions from the data based on similarity and dissimilarity constraints. Let $\{x_i\}_{i=1}^N$ be a set of $N$ points in $\mathcal{R}^d$. Given pairs of similar points $S$ and pairs of dissimilar points $D$, the aim of ITML is to learn an SPD matrix $M$ such that the Mahalanobis distance parametrized by $M$ is below a given threshold $l$ for similar pairs of points and above a given threshold $u$ for dissimilar pairs of points.\\[10pt]
Let $D_{ld}$ denote the LogDet divergence between SPD matrices defined as
\begin{equation}
D_{ld}(P, Q) = \text{trace}(PQ^{-1}) - \text{log det}(PQ^{-1}) - n;\  P,\  Q \in S_n^{++}.
\end{equation}
ITML formulates the Mahalanobis matrix learning as the following optimization problem:
\begin{equation}
\begin{aligned}
& \underset{M \succ 0,\ \zeta}{\text{minimize}}\ \ D_{ld} (M, M_0) + \gamma D_{ld}(\text{diag}(\zeta), \text{diag}(\zeta_0))\\
& \text{subject to}\ \ (x_i - x_j)^\top M (x_i - x_j) \leq \zeta_{c(i,j)},\ \forall (i,j) \in S\\
& \hspace{59pt} (x_i - x_j)^\top M (x_i - x_j) \geq \zeta_{c(i,j)},\ \forall (i,j) \in D,
\end{aligned}
\end{equation}
where $c(i,j)$ denotes the index of the $(i,j)-$th constraint, $\zeta$ is the vector of variables $\zeta_{c(i,j)}$, $\zeta_0$ is a vector whose components equal $l$ for similarity constraints and $u$ for dissimilarity constraints, $M_0$ is an SPD matrix that captures the prior knowledge about $M$, and $\gamma$ is a parameter controlling the tradeoff between satisfying the constraints and minimizing $D_{ld}(M, M_0)$. This optimization problem can be solved efficiently using Bregman iterations. In this work, we use the publicly available ITML code provided by the authors of [7].

\textbf{ITML parameters:} We need to specify the values for the following parameters while using ITML: $M_0,\ \gamma,\ l,\ u$. We choose the constraint thresholds $l$ and $u$ as the $a^{th}$ and $b^{th}$ percentiles of the observed distribution of distances between pairs of points within the training dataset. Hence, the parameters for the ITML algorithm are $M_0,\ \gamma,\ a$ and $b$.
\begin{table}
\small
\center
\caption{SPD matrix distances and their properties}
\renewcommand{\arraystretch}{1.8}
\begin{tabular}{| m{2.3cm} | c | c | m{1.7cm} | c |}\hline
Distance  & Formula & Symmetric & Triangle inequality & Geodesic \\ \hline
Frobenius & $\|P_1 - P_2\|_F$ & Yes & Yes & No\\ \hline
%Power-Euclidean [13] & $\frac{1}{\alpha}\|P_1^\alpha - P_2^\alpha \|_F$ & Yes & Yes & No\\ \hline
Cholesky-Frobenius [13] & $\|\text{Chol}(P_1) - \text{Chol}(P_2)\|_F$ & Yes & Yes & No\\ \hline
J-divergence [12] & $\frac{1}{2}\sqrt{\text{trace}(P_1P_2^{-1} + P_2P_1^{-1}) - 2n}$ & Yes & No & No\\ \hline
Jensen-Bregman LogDet Divergence[11] & $\sqrt{\text{log det} \left(\frac{P_1 + P_2}{2}\right) - \frac{1}{2} \text{log det}\left(P_1P_2\right)}$ & Yes & No & No\\ \hline
Affine-invariant [1] & $\|\text{log}\left(P_1^{-1/2}P_2P_1^{-1/2}\right)\|_F$ & Yes & Yes & Yes\\ \hline
Log-Frobenius [6] & $\| \text{log}(P_1) - \text{log}(P_2) \|_F$ & Yes & Yes & Yes\\ \hline
\end{tabular}
\caption{SPD matrix distances and their properties}
\begin{tabular}{| m{3cm} | m{1.3cm} | m{1.5cm} | m{1.5cm} | m{1.5cm} | m{1.5cm} |}
\hline
Distance & Distance from $S_n^+$ & Affine invariance & Scale invariance & Rotation invariance & Inversion invariance\\ \hline
Frobenius & Finite & No & No & Yes & No\\ \hline
%Power Euclidean [13] & Finite & No & No & Yes & No\\ \hline
Cholesky-Frobenius [13] & Finite & No & No & No & No\\ \hline
J-divergence [12] & Infinite & Yes & Yes & Yes & Yes\\ \hline
Jensen-Bregman LogDet Divergence[11] & Infinite & Yes & Yes & Yes & Yes\\ \hline
Affine-invariant [1] & Infinite & Yes & Yes & Yes & Yes\\ \hline
Log-Frobenius [6] & Infinite & No & Yes & Yes & Yes\\ \hline
\end{tabular}
\end{table}
\section{Log-Euclidean Riemannian metric learning}
The log-Euclidean framework [6] proposed by Arsigny \textit{et.\ al.}\ defines a class of Riemannian metrics called log-Euclidean metrics. The geodesic distances associated with log-Euclidean metrics are called log-Euclidean distances. Let $\odot$ be an operation on SPD matrices defined as $P_1 \odot P_2 = \text{exp}\left(\text{log}(P_1\right) + \text{log}(P_2))$. We have the following result based on the log-Euclidean framework introduced in [6]:\\[10pt]
\textbf{Result 4.1:} Any inner product $\left\langle\ ,\ \right\rangle$ defined on $\mathcal{T}_IS_n^{++}  = \{\text{log}(P)\ |\ P \in S_n^{++} \} = S_n$ extended to the Lie group $\left(S_n^{++}, \odot \right)$ by left- or right- multiplication is a bi-invariant Riemannian metric. The corresponding geodesic distance between $P_1 \in S_n^{++}$ and $P_2 \in S_n^{++}$ is given by
\begin{equation}
d(P_1, P_2) = \| \text{mlog}_I(P_1) - \text{mlog}_I(P_2) \|_I = \| \text{log}(P_1) - \text{log}(P_2) \|_I,
\end{equation}
where $\|\ \|_I$ is the norm induced by $\left\langle\ ,\ \right\rangle$. Note that here $\text{mlog}_I$ is the inverse-exponential map at the identity matrix which is equal to the usual matrix logarithm in this case.\\[10pt]
The set of all $n \times n$ symmetric matrices form a vector space of dimension $d = \frac{n(n+1)}{2}$. Let $vec(P)$ denote the column vector form of the upper triangular part of a matrix $P$. This $vec()$ operation provides a $d$ dimensional vector representation for $S_n$. Let  $\left\langle\ ,\ \right\rangle$ be an inner product defined on the vector space $S_n$ and $M \in S_d^{++}$ be the corresponding matrix of inner products between the $d$ basis vectors corresponding to $vec()$ representation. Note that $\left\langle\ ,\ \right\rangle$ is uniquely characterized by $M$. The distance between two matrices $P_1 \in S_n$ and $P_2 \in S_n$ induced by this inner product is given by
\begin{equation}
d(P_1, P_2) = \left(vec(P_1) - vec(P_2)\right)^\top M \left(vec(P_1) - vec(P_2)\right).
\end{equation}
\textbf{Result 4.2:} Let $M \in S_d^{++}$ , where $d = \frac{n(n+1)}{2}$. Then, $M$ defines a unique inner product denoted by $\left\langle\ ,\ \right\rangle_M$  on $\mathcal{T}_IS_n^{++} = \{\text{log}(P)\ |\ P \in S_n^{++} \} = S_n$. This inner product $\left\langle\ ,\ \right\rangle_M$ also defines a log-Euclidean Riemannian metric which can be obtained by simply extending $\left\langle\ ,\ \right\rangle_M$  to the Lie group  $\left(S_n^{++}, \odot \right)$ by left- or right- multiplication. The corresponding log-Euclidean geodesic distance between $P_1 \in S_n^{++}$ and $P_2 \in S_n^{++}$ is given by
\begin{equation}
d_M(P_1, P_2) = \left(vec(\text{log}(P_1)) - vec(\text{log}(P_2))\right)^\top M \left(vec(\text{log}(P_1)) - vec(\text{log}(P_2))\right).
\end{equation}
The above result follows directly from result 4.1. Result 4.2 says that any Mahalanobis distance defined in the vector space $\{vec(\text{log}(P))\ |\ P \in S_n^{++} \}$ is a geodesic distance on $S_n^{++}$ and the corresponding Riemannian metric is uniquely defined by the Mahalanobis matrix $M$. Hence, we can learn Riemannian metrics/geodesic distances for $S_n^{++}$ from the data by learning Mahalanobis distance functions in the vector space $\{vec(\text{log}(P))\ |\ P \in S_n^{++} \}$. Table 3 summarizes our  approach for leaning geodesic distances on $S_n^{++}$. In this work, we use ITML technique for Mahalanobis distance learning.
\begin{table}
\center
\caption{Algorithm for learning geodesic distances on $S_n^{++}$}
\renewcommand{\arraystretch}{1.2}
\begin{tabular}{|m{11.2cm}|}\hline
\textbf{Input:} $\{P_i \in S_n^{++}\}_{i=1}^N$\\ \hline
\textbf{for}\ \ $ i = 1$ to $N$\\
\hspace{20pt} $v_i = vec(\text{log}(P_i))$\\
\textbf{end}\\
\hline
Learn a Mahalanobis distance function using $\{v_i\}_{i=1}^N.$ This gives a Mahalanobis matrix $M \in S^{++}_d$, where $d = \frac{n(n+1)}{2}$.\\
\hline
\textbf{Output:} Geodesic distance between $P_1$ and $P_2$:\\
$\left(vec(\text{log}(P_1)) - vec(\text{log}(P_2))\right)^\top M \left(vec(\text{log}(P_1)) - vec(\text{log}(P_2))\right)$. \\
\hline
\end{tabular}
\end{table}
\newpage
\section{Experiments}
In the section, we evaluate the performance of the proposed Riemannian metric/geodesic distance learning approach on two applications: (i) Face matching using Labeled Faces in the Wild (LFW) dataset and (ii) Semi-supervised clustering using ETH80 dataset.
\subsection{Face matching using LFW face dataset}
In this experiment our aim is to predict whether a given pair of face images correspond to the same person or not.\\[5pt]
\textbf{Dataset:} The LFW dataset [9] is a collection of face photographs designed for studying the problem of unconstrained face recognition. This dataset consists of 13233 labeled face images of 1680 subjects collected from the web.  This dataset consists of two subsets:
\begin{itemize}
\item \textbf{Development subset:} The development subset consists of 2200 training image pairs, where 1100 are similar pairs and 1100 are dissimilar pairs, and 1000 test image pairs, where 500 are similar pairs and 500 are dissimilar pairs. An image pair is said to be similar if both the images correspond to the same person and dissimilar if they correspond to different persons.
\item \textbf{Evaluation subset:} The evaluation subset consists of 3000 similar image pairs and 3000 dissimilar image pairs. It is further divided into 10 subsets each of which consists of 300 similar pairs and 300 dissimilar pairs.
\end{itemize}
All the image pairs were generated by randomly selecting images from the 13233 images in the dataset. The development subset is meant for model and parameter selection. The evaluation subset should be used only once for final training and testing. To avoid overfitting, the image pairs in the development subset were chosen to be different from the image pairs in the evaluation subset. 
\subsubsection{Feature extraction}
We crop the face region in each image and resize it to a $64 \times 64$ image. Following [3], we convert each pixel in an image into a 9-dimensional feature vector given by
\begin{equation*}
\tiny
\left[x,\ y,\ R(x,y),\ G(x,y),\ B(x,y),\ \left| \frac{\partial W(x,y)}{\partial x} \right|,\ \left|\frac{\partial W(x,y)}{\partial y}\right|,\ \left|\frac{\partial^2 W(x,y)}{\partial x^2}\right|,\ \left|\frac{\partial^2 W(x,y)}{\partial y^2} \right| \right]^\top,
\end{equation*}
where $x,\ y$ are the column and row coordinates respectively, $R, G$ and $B$ are the color coordinates and $W$ is the grayscale image. We use the $9 \times 9$ covariance matrix of the feature vectors to represent the image.
\subsubsection{Experimental protocol}
Following the standard experimental protocol for this dataset, we use the development set for selecting the parameters of ITML and then use the evaluation set only once for final training and testing. Following steps summarize our experimental procedure:
\begin{itemize}
\item \textbf{Parameter selection:} We train the ITML algorithm using the 2200 training pairs of the development subset and then test it on the 1000 test pairs of the development subset. We select the ITML parameters that give the best test accuracy.
\item \textbf{Final training and testing:} The evaluation set consists of 10 splits and we perform 10-fold cross-validation. In each fold, we use 9 splits (2700 similar pairs and 2700 dissimilar pairs) for training ITML and 1 split (300 similar pairs and 300 dissimilar pairs) for testing. For training ITML, we use the parameters that were selected in the previous step. Since our task is face matching, we need to threshold the learned distance function. In each fold, we find the threshold that gives best training accuracy and use the same threshold for test image pairs.
\end{itemize}
\subsubsection{Comparative methods}
We compare the performance of the proposed log-Euclidean metric learning approach with the following approaches:
\begin{itemize}
\item Directly use any of the following distances for matching:
\begin{itemize}
\item Frobenius,  Cholesky-Frobenius, J-divergence, Jensen-Bregman LogDet divergence, Affine-invariant and Log-Frobenius.
\end{itemize}
\item Use ITML directly with the covariance matrices by treating them as elements of the Euclidean space of symmetric matrices.
\item Use ITML with the lower triangular matrix obtained by Cholesky decomposition.
\end{itemize}
In all these methods the distance threshold is obtained in each fold independently based on the training data.
\subsubsection{Parameters}
The following parameter values were used for ITML:
\begin{itemize}
\item $M_0 = I,\ \gamma = 10^{3.5},\ a = 5,\ b = 95.$
\end{itemize}
These parameters were selected using the development subset of the dataset.
\begin{table}
\small
\center
\caption{Prediction accuracy on LFW dataset using various SPD matrix distances}
\renewcommand{\arraystretch}{1.3}
\begin{tabular}{| m{1.4cm} | m{1.7cm} |  m{1.9cm}| m{1.8cm} | m{2.4cm} | m{1.5cm} |}\hline
Frobenius & Cholesky-Frobenius & Log-Frobenius & J-divergence & Jenson-Bregman LogDet divergence & Affine invariant\\ \hline
\ \ 53.77 &\ \  56.62 &\ \ \ \ \ 60.43 &\ \ \ \ 60.92 &\ \ \ \ \ \ \ \  61.62 &\ \ \ \ 61.15\\ \hline
\end{tabular}
\caption{Prediction accuracy on LFW dataset using distance learning}
\begin{tabular}{| m{1.3cm} | m{0.9cm} | m{1.4cm} | m{1.3cm} | m{0.9cm} | m{1.4cm} | m{1.3cm} | m{0.9cm} | m{1.4cm} |}
\hline
\multicolumn{3}{|c|}{Covariance matrices} & \multicolumn{3}{c|}{Cholesky decompositions} & \multicolumn{3}{c|}{Log-Euclidean}\\ \hline
Frobenius & ITML & ITML gain & Frobenius & ITML & ITML gain & Frobenius & ITML & ITML gain \\
\hline
\ \ \ 53.77 & 57.58 &\ \ \ \ \  3.81 &\ \ \ 56.62 & 63.53 &\ \ \ \ \  6.91 &\ \ \ 60.43 & \textbf{69.37} &\ \ \ \ \  8.94   \\
\hline
\end{tabular}
\end{table}
\subsubsection{Results}
Tables 4 and 5 summarize the prediction results for various approaches on the LFW data set. We can draw the following conclusions from these results:
\begin{itemize}
\item The proposed Riemannian metric/geodesic distance learning approach outperforms the other approaches for comparing covariance matrices.
\item The log-Euclidean geodesic distance learned from the data performs much better than the standard log-Frobenius distance.
\item Distance learning with original covariance matrices or Cholesky decompositions performs poorly compared to distance learning in the logarithm domain.
\end{itemize}
\subsection{Semi-supervised clustering using ETH80 object dataset}
In this experiment, we are interested in clustering the images in the ETH80 dataset into different object categories.
\subsubsection{Dataset}
The ETH80 object dataset [10] consists of $256 \times 256$ images of 8 object categories with each category including 10 different object instances. Each object instance has 41 images captured under different views. So, each object category has 410 images resulting in a total of 3280 images.
\subsubsection{Feature extraction}
We convert each pixel in an image into a 9-dimensional feature vector given by
\begin{equation*}
\tiny
\left[x,\ y,\ R(x,y),\ G(x,y),\ B(x,y),\ \left| \frac{\partial W(x,y)}{\partial x} \right|,\ \left|\frac{\partial W(x,y)}{\partial y}\right|,\ \left|\frac{\partial^2 W(x,y)}{\partial x^2}\right|,\ \left|\frac{\partial^2 W(x,y)}{\partial y^2} \right| \right]^\top,
\end{equation*}
where $x,\ y$ are the column and row coordinates respectively, $R, G$ and $B$ are the color coordinates and $W$ is the grayscale image. We compute the $9 \times 9$ covariance matrix of the feature vectors over the entire image and use it to represent the image.
\subsubsection{Experimental protocol and parameters}
For every object category, we randomly select 4 images from each instance for training. Hence, we use 40 samples from each object category for training, resulting in a total of 320 training images. From each pair of training images, we generate either a similarity constraint or a dissimilarity constraints based on their category labels. We use all such constraints in learning the Mahalanobis distance function. Once we learn the Mahalanobis distance function, we use it for clustering the entire dataset of 3280 images.\\[10pt]
We repeat the above procedure 5 times and report the average clustering accuracy. In each run, we select the value of ITML parameter $\gamma$ using two fold cross-validation on the training data. We use the following values for other ITML parameters in all the 5 runs: $M_0 = I,\ a = 5,\ b = 95$.\\[10pt]
We use K-means algorithm for clustering. To handle the local-optimum issue, we run K-means with 20 different random initializations and select the clustering result corresponding to the minimum K-means cost value.
\subsubsection{Comparative methods}
We compare the performance of the proposed log-Euclidean metric learning approach with the following approaches:
\begin{itemize}
\item \textbf{Unsupervised:} Directly perform K-means clustering using any of the following distances: Frobenius, Cholesky-Frobenius and Log-Frobenius.
\item Use ITML directly with the covariance matrices by treating them as elements of the Euclidean space of symmetric matrices.
\item Use ITML with the lower triangular matrix obtained by Cholesky decomposition.
\end{itemize}
Computation of mean doesn't have a closed form solution in the case of J-divergence or Jensen-Bregman LogDet divergence or Affine-invariant distance. Hence, we need to use some optimization procedure for computing the mean. This makes K-means algorithm highly computational. Hence, we do not use these distances for comparison in this work.
\subsubsection{Results}
Table 6 summarizes the clustering results for various approaches on the ETH80 dataset. We can draw the following conclusions from these results:
\begin{itemize}
\item The proposed Riemannian metric/geodesic distance learning approach performs better than other approaches for clustering SPD matrices.
\item The log-Euclidean geodesic distance learned from the data performs much better than the standard log-Frobenius distance.
\item Distance learning with original covariance matrices or Cholesky decompositions performs poorly compared to distance learning in the logarithm domain.
\end{itemize}
\section{Conclusion}
In this work, we have explored the idea of data-driven Riemannian metrics or geodesic distances. Based on the log-Euclidean framework [6], we have shown how geodesic distance functions can be learned for $S_n^{++}$ by simply learning Mahalanobis distance functions in the logarithm domain. We have conducted experiments using face and object data sets. The face matching and semi-supervised object categorization results clearly show that the learned log-Euclidean geodesic distance performs much better than other distances.
\begin{table}
\center
\caption{Clustering accuracy on ETH80 dataset}
\begin{tabular}{| m{1.4cm} | m{0.9cm} | m{1.4cm} | m{1.4cm} | m{0.9cm} | m{1.4cm} | m{1.4cm} | m{0.9cm} | m{1.4cm} |}
\hline
\multicolumn{3}{|c|}{Covariance matrices} & \multicolumn{3}{c|}{Cholesky decompositions} & \multicolumn{3}{c|}{Log-Euclidean}\\ \hline
Frobenius & ITML & ITML gain & Frobenius & ITML & ITML gain & Frobenius & ITML & ITML gain \\
\hline
\ \ \ 35.58& 70.50 &\ \ \ \ \ 34.92 &\ \ \ 51.13 & 70.36 &\ \ \ \ \ 19.24 &\ \ \ 55.70 & \textbf{73.79} &\ \ \ \ \  18.09   \\
\hline
\end{tabular}
\end{table}
\newpage
\section*{References}
\begin{enumerate}
\item X.\ Pennec, P.\ Fillard, and N.\ Ayache, \lq\lq A Riemannian Framework for Tensor Computing\rq\rq, \textit{IJCV}, 2006.\\
\item A.\ Goh and R.\ Vidal, \lq\lq Clustering and Dimensionality Reduction on Riemannian Manifolds\rq\rq, \textit{In CVPR}, 2008.\\
\item O.\ Tuzel, F.\ Porikli, and P.\ Meer, \lq\lq Region Covariance: A Fast Descriptor for Detection and Classification\rq\rq, \textit{In ECCV}, 2006.\\
\item O.\ Tuzel, F.\ Porikli, and P.\ Meer, \lq\lq Pedestrian Detection via Classification on Riemannian Manifolds\rq\rq, \textit{PAMI}, 2008.\\
\item M.\ Harandi, C.\ Sanderson, R.\ Hartley, and B.\ Lovel, \lq\lq Sparse Coding and Dictionary Learning for Symmetric Positive Definite Matrices: A Kernel Approach\rq\rq, \textit{In ECCV}, 2012.\\
\item V.\ Arsigny, P.\ Fillard, X.\ Pennec, and N.\ Ayache, \lq\lq Log-Euclidean Metrics for Fast and Simple Calculus on Diffusion Tensors\rq\rq, \textit{Magnetic Resonance in Medicine}, 2006.\\
\item J.\ V.\ Davis, B.\ Kulis, P.\ Jain, S.\ Sra, and I.\ S.\ Dhillon, \lq\lq Information-Theoretic Metric Learning \rq\rq, \textit{In ICML}, 2007.\\
\item K.\ Q.\ Weinberger and L.\ K.\ Saul, \lq\lq Distance Metric Learning for Large Margin Nearest Neighbor Classification\rq\rq, \textit{JMLR}, 2009.\\
\item G.\ B.\ Huang, M.\ Ramesh, T.\ Berg, and E.\ L.-Miller, \lq\lq Labeled Faces in the Wild: A Database for Studying Face Recognition in Unconstrained Environments\rq\rq, \textit{University of Massachusetts, Amherst, Technical Report 07-49}, October, 2007.\\
\item B.\ Leibe and B.\ Schiele, \lq\lq Analyzing Appearance and Contour Based Methods for Object Categorization\rq\rq, \textit{In  CVPR}, 2003.\\
\item A.\ Cherian, S.\ Sra, A.\ Banerjee, and N.\ Papanikolopoulos, \lq\lq Efficient Similarity Search for Covariance Matrices via the Jensen-Bregman LogDet Divergence\rq\rq, \textit{In ICCV}, 2011.\\
\item Z.\ Wang and B.\ C.\ Vemuri, \lq\lq An Affine Invariant Tensor Dissimilarity Measure and its Applications to Tensor-valued Image Segmentation\rq\rq, \textit{In CVPR}, 2004.\\
\item I.\ L.\ Dryden, A.\ Koloydenko, and D.\ Zhou, \lq\lq Non-Euclidean Statistics for Covariance Matrices, with Applications to Diffusion Tensor Imaging\rq\rq, \textit{The Annals of Applied Statistics}, 2009.\\
\end{enumerate}
\end{document}